\documentclass[10pt,twocolumn,letterpaper]{article}

\usepackage[utf8]{inputenc} 
\usepackage[T1]{fontenc}    

\usepackage{cvpr}
\usepackage{times}
\usepackage{epsfig}
\usepackage{graphicx}
\usepackage{amsmath}
\usepackage{amssymb}
\usepackage{float}
\usepackage{stfloats}
\usepackage{adjustbox}
\usepackage{pifont}
\usepackage{mathtools,cases}
\usepackage{empheq}
\usepackage{bm}
\usepackage{makecell}
\def\eg{\textit{e.g.}~}
\def\ie{\textit{i.e.}~}

\usepackage[pagebackref=true,breaklinks=true,letterpaper=true,colorlinks,bookmarks=false]{hyperref}

\cvprfinalcopy 

\def\cvprPaperID{****} 

\ifcvprfinal\fi
\begin{document}

\title{Disentangling Physical Dynamics from Unknown Factors for Unsupervised Video Prediction}

\author{
  Vincent Le Guen  $^{1,2}$,  Nicolas Thome $^2$ \\
  $^1$ EDF R\&D, Chatou, France \\
  $^2$ CEDRIC, Conservatoire National des Arts et Métiers, Paris, France 
}

\maketitle

\begin{abstract}
Leveraging physical knowledge described by partial differential equations (PDEs) is an appealing way to improve unsupervised video prediction methods. Since physics is too restrictive for describing the full visual content of generic videos, we introduce PhyDNet, a two-branch deep architecture, which explicitly disentangles PDE dynamics from unknown complementary information. A second contribution is to propose a new  recurrent physical cell (PhyCell), inspired from data assimilation techniques, for performing PDE-constrained prediction in latent space. Extensive experiments conducted on four various datasets show the ability of PhyDNet to outperform state-of-the-art methods. Ablation studies also highlight the important gain brought out by both disentanglement and PDE-constrained prediction. Finally, we show that PhyDNet presents interesting features for dealing with  missing data and long-term forecasting.
\end{abstract}

\section{Introduction}
\label{sec:intro}
Video forecasting consists in predicting the future content of a video conditioned on previous frames. This is of crucial importance in various contexts, such as weather forecasting \cite{xingjian2015convolutional}, autonomous driving \cite{kwon2019predicting}, reinforcement learning \cite{oh2015action}, robotics \cite{finn2016unsupervised}, or action recognition \cite{liu2017video}. 
In this work, we focus on unsupervised video prediction, where the absence of semantic labels to drive predictions exacerbates the challenges of the task. 
In this context, a key problem is to design video prediction methods able to represent the complex dynamics underlying raw data.

State-of-the-art methods for training such complex dynamical models currently rely on deep learning, with specific architectural choices based on 2D/3D convolutional~\cite{mathieu2015deep,vondrick2016generating} or recurrent neural networks~\cite{wang2017predrnn,wang2018predrnn++,wang2019memory}. 

To improve predictions, recent methods use adversarial training \cite{mathieu2015deep,vondrick2016generating,kwon2019predicting}, stochastic models \cite{castrejon2019improved,minderer2019unsupervised}, constraint predictions by using geometric knowledge \cite{finn2016unsupervised,jia2016dynamic,xue2016visual} or by disentangling factors of variation \cite{villegas2017decomposing,tulyakov2018mocogan,denton2017unsupervised,hsieh2018learning}.

\begin{figure}
    \centering
    \includegraphics[width=\columnwidth]{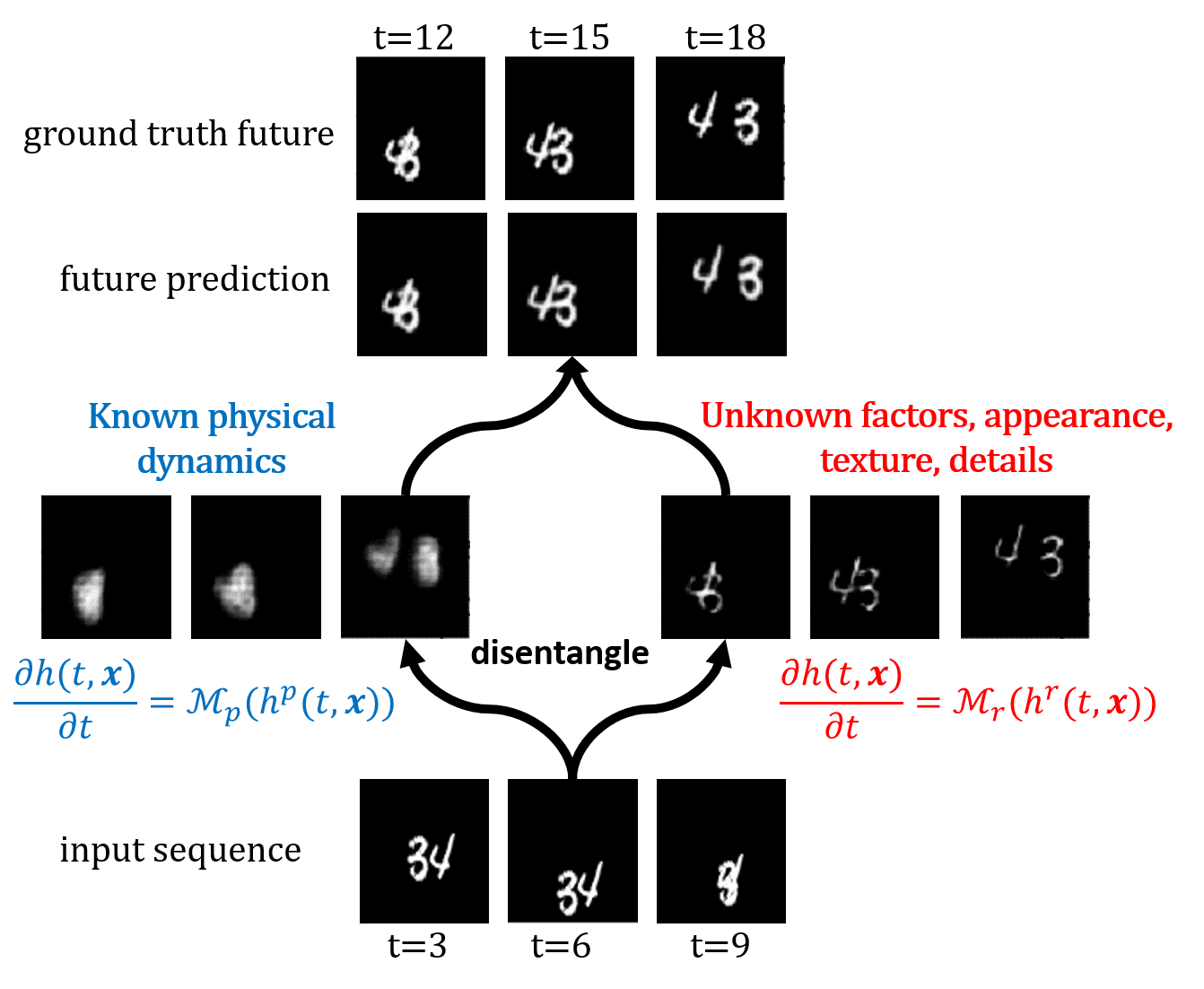}
        \caption{PhyDNet is a deep model mapping an input video into a latent space $\bm{\mathcal{H}}$, from which future frame prediction can be accurately performed. PhyDNet learns $\bm{\mathcal{H}}$ in an unsupervised manner, such that physical dynamics and unknown factors necessary for prediction, \eg appearance, details, texture, are disentangled. \vspace{-0.1cm}} 
    \label{fig:fig1}
\end{figure}

Another appealing way to model the video dynamics is to exploit prior physical knowledge, \eg formalized by partial differential equations (PDEs) \cite{de2017deep,seo2019differentiable}. Recently, interesting connections between residual networks and PDEs have been drawn \cite{weinan2017proposal,lu2018beyond,chen2018neural}, enabling to design physically-constrained machine learning frameworks~\cite{raissi2018deep,de2017deep,seo2019differentiable,rudy2017data}. 
These approaches are very successful for modeling complex natural phenomena, \eg climate, when the underlying dynamics is well described by the physical equations in the input space~\cite{raissi2018deep,rudy2017data,long2018pde}. However, such assumption is rarely fulfilled in the pixel space for predicting generalist videos.
  
 In this work,  we introduce PhyDNet, a deep model dedicated to perform accurate future frame predictions from generalist videos. In such a context, physical laws do not apply in the input pixel space ; the goal of PhyDNet is to learn a semantic latent space $\bm{\mathcal{H}}$ in which they do, and are disentangled from other factors of variation required to perform future prediction.
Prediction results of PhyDNet 
  when trained on Moving MNIST~\cite{srivastava2015unsupervised} are shown in Figure \ref{fig:fig1}. The left branch represents the physical dynamics in $\bm{\mathcal{H}}$ ; when decoded in the image space, we can see that the corresponding features encode approximate segmentation masks predicting digit positions on subsequent frames. 
  On the other hand, the right branch extracts residual information required for prediction, here the precise appearance of the two digits. Combining both representations eventually makes accurate prediction successful.
  
Our contributions to the unsupervised video prediction problem with PhyDNet can be summarized as follows: 
\begin{itemize}
\item We introduce a global sequence to sequence two-branch deep model (section~\ref{sec:3.1}) dedicated to jointly learn the latent space $\bm{\mathcal{H}}$ and to disentangle physical dynamics from residual information, the latter being modeled by a data-driven (ConvLSTM~\cite{xingjian2015convolutional}) method. \vspace{-0.05cm}

\item Physical dynamics is modeled by a new recurrent physical cell, PhyCell (section~\ref{section:phycell}), discretizing a broad class of PDEs in $\bm{\mathcal{H}}$. 
PhyCell is based on a prediction-correction paradigm inspired from the data assimilation community \cite{asch2016data},~enabling robust training with missing data and for long-term forecasting. \vspace{-0.05cm}

\item Experiments (section~\ref{section4}) reveal that PhyDNet outperforms state-of-the-art methods on four generalist datasets: this is, as far as we know, the first physically-constrained model able to show such capabilities. We highlight the importance of both disentanglement and physical prediction for optimal performances.

\end{itemize}

\section{Related work}
\label{sec:sota}

We review here related multi-step video prediction approaches dedicated to long-term forecasting. We also focus on unsupervised training, \ie only using input video data and without manual supervision based on semantic labels. \vspace{-0.2cm}

\begin{figure*}
    \centering
    \begin{tabular}{cc}
        \includegraphics[width=6cm]{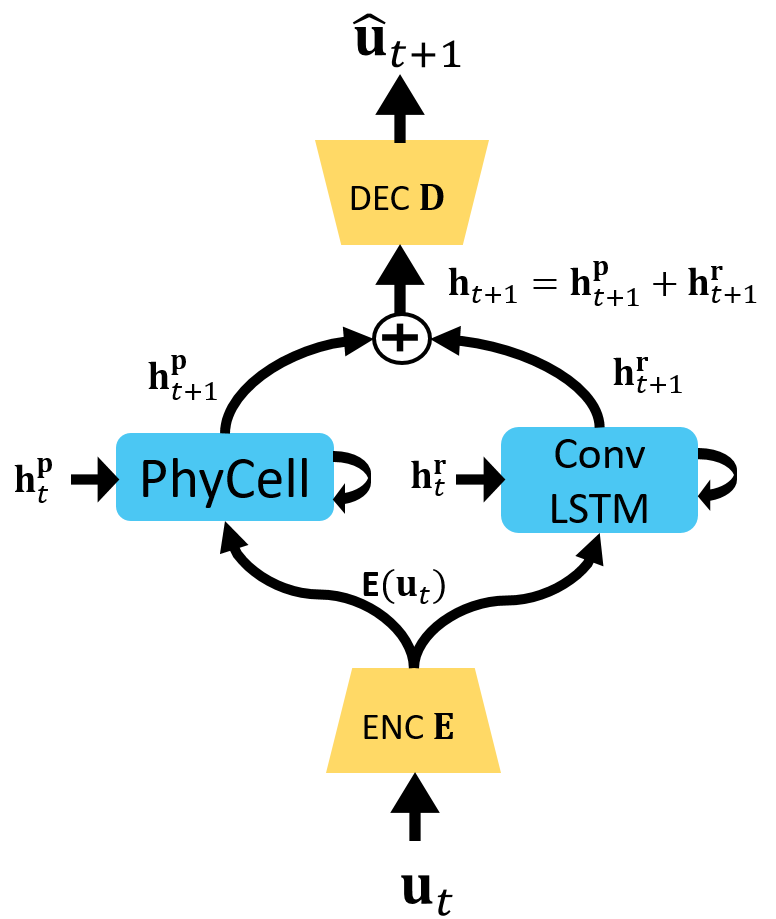} & \includegraphics[width=11cm]{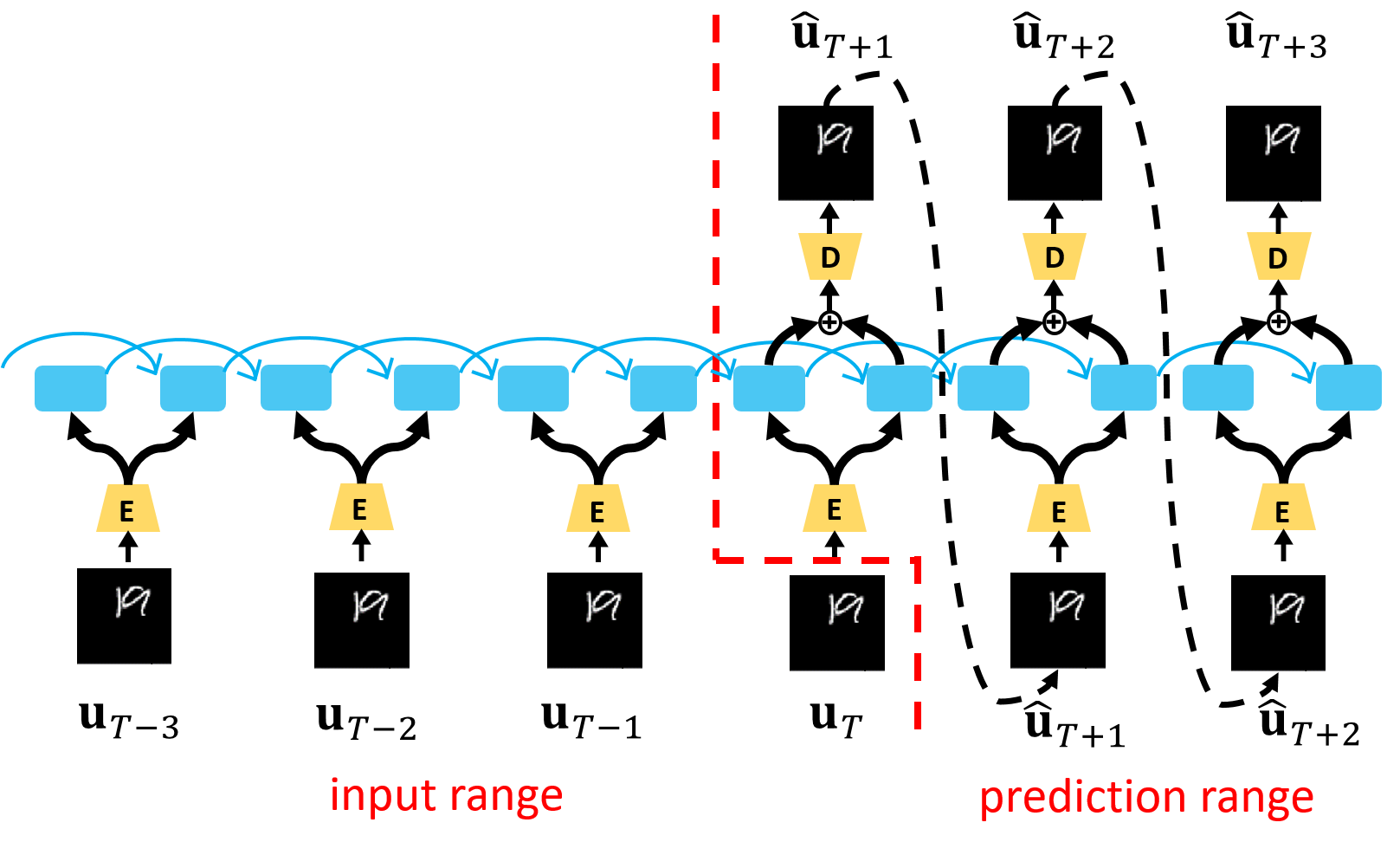}  \\
        \textbf{(a) PhyDNet disentangling recurrent bloc} & \textbf{(b) Global seq2seq architecture} \vspace{0.2cm}
    \end{tabular}{}
    \caption{\textbf{Proposed PhyDNet deep model for video forecasting.} a) The core of PhyDNet is a recurrent block projecting input images $\mathbf{u_t}$ into a latent space $\bm{\mathcal{H}}$, where two recurrent neural networks disentangle physical dynamics (PhyCell, section \ref{section:phycell}) from residual information (ConvLSTM). Learned physical $\mathbf{h}^{\mathbf{p}}_{t+1}$ and residual $\mathbf{h}^{\mathbf{r}}_{t+1}$ representations are summed before decoding to predict the future image $\hat{\mathbf{u}}_{t+1}$. b) Unfolded in time, PhyDNet forms a sequence to sequence (seq2seq) architecture suited for multi-step video prediction. Dotted arrows mean that  predictions are reinjected as next input only for the ConvLSTM branch, and not for PhyCell, as explained in section \ref{sec:training}.}
    \label{fig:fig2}
\end{figure*}

\paragraph{Deep video prediction} 
Deep neural networks have recently achieved state-of-the-art performances for data-driven video prediction. Seminal works include the application of sequence to sequence LSTM  or Convolutional variants~\cite{srivastava2015unsupervised,xingjian2015convolutional}, adopted in many studies \cite{finn2016unsupervised,lu2017flexible,xu2018structure}. Further works explore different architectural designs based on Recurrent Neural Networks (RNNs) \cite{wang2017predrnn,wang2018predrnn++,oliu2018folded,wang2019memory,wang2018eidetic} and 2D/3D ConvNets \cite{mathieu2015deep,vondrick2016generating,reda2018sdc,byeon2018contextvp}. Dedicated loss functions \cite{cuturi2017soft,leguen19} and Generative Adversarial Networks (GANs) have been investigated for sharper predictions \cite{mathieu2015deep,vondrick2016generating,kwon2019predicting}. However, the problem of conditioning GANs with prior information, such as physical models, remains an open question. 

To constrain the challenging generation of high dimensional images, several methods rather predict geometric transformations between frames \cite{finn2016unsupervised,jia2016dynamic,xue2016visual} or use optical flow \cite{patraucean2015spatio,luo2017unsupervised,liu2017video,liang2017dual,li2018flow}. This is very effective for short-term prediction, but degrades quickly when the video content evolves, where more complex models and memory about dynamics are required. 

A promising line of work consists in disentangling independent factors of variations in order to apply the prediction model on lower-dimensional representations. A few approaches explicitly model interactions between objects inferred from an observed scene \cite{eslami2016attend,kosiorek2018sequential,ye2019compositional}. Relational reasoning, often implemented with graphs \cite{battaglia2016interaction,kipf2018neural,sanchez2018graph,palm2018recurrent,van2018relational}, can account for basic physical laws, \eg drift, gravity, spring \cite{watters2017visual,wu2017learning,mrowca2018flexible}. However, these methods are object-centric, only evaluate on controlled settings and are not suited for general real-world video forecasting.
Other disentangling approaches factorize the video into independent components \cite{villegas2017decomposing,tulyakov2018mocogan,denton2017unsupervised,hsieh2018learning,gao2019disentangling}. Several disentanglement criteria are used, such as content/motion \cite{villegas2017decomposing} or deterministic/stochastic \cite{denton2017unsupervised}. In specific contexts, the prediction space can be structured using additional information, \eg with human pose \cite{villegas2017learning,walker2017pose} or key points \cite{minderer2019unsupervised}, which imposes a severe overhead on the annotation budget. \vspace{-0.2cm} 

\paragraph{Physics and PDEs}
Exploiting prior physical knowledge is another appealing way to improve prediction models. Earlier attempts for data-driven PDE discovery include sparse regression of potential differential terms \cite{brunton2016discovering,rudy2017data,schaeffer2017learning} or  neural networks  approximating the solution and response function of PDEs \cite{raissi2017physics,raissi2018deep,seo2019differentiable}. Several approaches are dedicated to a specific PDE, \eg  advection-diffusion in~\cite{de2017deep}.
~Based on the connection between numerical schemes for solving PDEs (\eg Euler, Runge-Kutta) and residual neural networks \cite{weinan2017proposal,lu2018beyond,chen2018neural,zhu2018convolutional}, several specific architectures were designed for predicting and identifying dynamical systems \cite{fablet2018bilinear,long2018pde,qin2019data}. 
PDE-Net \cite{long2018pde,long2019pde} discretizes a broad class of PDEs by approximating partial derivatives with convolutions. 
Although these works leverage physical knowledge, they either suppose physics behind data to be explicitly known or are limited to a fully visible state, which is rarely the case for general video forecasting. \vspace{-0.2cm}

\paragraph{Deep Kalman filters}
To handle unobserved phenomena, state space models, in particular the Kalman filter \cite{kalman1960new}, have been recently integrated with deep learning, by modeling dynamics in learned latent space \cite{Krishnan2015DeepKF,watter2015embed,haarnoja2016backprop,fraccaro2017disentangled,becker2019recurrent}. The Kalman variational autoencoder \cite{fraccaro2017disentangled} separates state estimation in videos from dynamics with a linear gaussian state space model. The Recurrent Kalman Network \cite{becker2019recurrent} uses a factorized high dimensional latent space in which the linear Kalman updates are simplified and don't require computationally-heavy covariance matrix inversions. These methods inspired by the data assimilation community \cite{asch2016data,bocquet2019data} have advantages in missing data or long-term forecasting contexts due to their mechanisms decoupling latent dynamics and input assimilation. However, they assume simple latent dynamics (linear) and don't include any physical prior. 



\section{PhyDNet model for video forecasting}
\label{section3}

We introduce PhyDNet, a model dedicated to video prediction, which leverages physical knowledge on dynamics, and disentangles it from other unknown factors of variations necessary for accurate forecasting. To achieve this goal, we introduce a disentangling architecture (section~\ref{sec:3.1}), and a new physically-constrained recurrent cell (section~\ref{section:phycell}). \vspace{0.25cm} \\ 
\textbf{Problem statement} As discussed in introduction, physical laws do not apply 
at the pixel level for general video prediction tasks. However, we  assume that there exists a conceptual latent space $\bm{\mathcal{H}}$ in which physical dynamics and residual factors are linearly disentangled.
Formally, let us denote as  $\mathbf{u}= \mathbf{u}(t,\mathbf{x})$ the frame of a video sequence at time $t$, for spatial coordinates $\mathbf{x}=(x,y)$. $\mathbf{h}(t,\mathbf{x}) \in \bm{\bm{\mathcal{H}}}$ is the latent representation of the video up to time $t$, which decomposes as $\mathbf{h}=\mathbf{h^p}+\mathbf{h^r}$, where $\mathbf{h^p}$ (resp. $\mathbf{h^r}$) represents the physical (resp. residual) component of the disentanglement. The video evolution in the latent space $\bm{\bm{\mathcal{H}}}$ is thus governed by the following partial differential equation (PDE):
\begin{equation}
\!\!\!\dfrac{\partial \mathbf{h}(t,\mathbf{x})}{\partial t} \! = \!\frac{\partial \mathbf{h^p}}{\partial t} \!+\! \frac{\partial \mathbf{\mathbf{h^r}}}{\partial t} \!:=\! \bm{\mathcal{M}}_{p}(\mathbf{h^p},\mathbf{u}) + \bm{\mathcal{M}}_{r}(\mathbf{\mathbf{h^r}},\mathbf{u}) \!\!\!
\label{eq:eq1}
\end{equation}


$\bm{\mathcal{M}}_p(\mathbf{h^p},\mathbf{u})$ and $\bm{\mathcal{M}}_r(\mathbf{h^r},\mathbf{u})$ represent physical and residual dynamics in the latent space $\bm{\bm{\mathcal{H}}}$.

\subsection{PhyDNet disentangling architecture}
\label{sec:3.1}


The main goal of PhyDNet is to learn the mapping from input sequences to a latent space which approximates the disentangling properties formalized in Eq (\ref{eq:eq1}). 

To reach this objective, we introduce a recurrent bloc which is shown in Figure \ref{fig:fig2}(a). A video frame $\mathbf{u}_t$ at time $t$ is mapped by a deep convolutional encoder $\mathbf{E}$ into a latent space representing the targeted space $\bm{\mathcal{H}}$. $\mathbf{E}(\mathbf{u}_t)$ is then used as input for two parallel recurrent neural networks, incorporating this spatial representation into a dynamical model. 

The left branch in Figure \ref{fig:fig2}(a) models the latent representation $\mathbf{h^p}$ fulfilling the physical part of the PDE in Eq (\ref{eq:eq1}), \ie $\frac{\partial \mathbf{h^p}(t,\mathbf{x})}{\partial t} = \bm{\mathcal{M}}_{p}(\mathbf{h^p},\mathbf{u})$. This PDE is modeled 
by our recurrent physical cell described in section \ref{section:phycell}, PhyCell, which  
leads to the computation of $\mathbf{h}^{\mathbf{p}}_{t+1}$ from $\mathbf{E}(\mathbf{u}_t)$ and $\mathbf{h}_t^{\mathbf{p}}$. From the machine learning perspective, PhyCell leverages physical constraints to limit the number of model parameters, regularizes training and improves generalization.

The right branch in Figure \ref{fig:fig2}(a) models the latent representation $\mathbf{h^r}$ fulfilling the residual part of the PDE in Eq (\ref{eq:eq1}), \ie $\frac{\partial \mathbf{h^r}(t,\mathbf{x})}{\partial t} = \bm{\mathcal{M}}_{r}(\mathbf{h^r},\mathbf{u})$. Inspired by wavelet decomposition \cite{mallat1999wavelet} and recent semi-supervised works \cite{robert2018hybridnet}, this part of the PDE corresponds to unknown phenomena, which do not correspond to any prior model, and is therefore entirely learned from data. We use a generic recurrent neural network for this task, \eg ConvLSTM~\cite{xingjian2015convolutional} for videos, which computes $\mathbf{h}_{t+1}^{\mathbf{r}}$ from $\mathbf{E}(\mathbf{u}_t)$ and $\mathbf{h}_{t}^{\mathbf{r}}$. 

$\mathbf{h}_{t+1}=\mathbf{h}_{t+1}^{\mathbf{p}} +\mathbf{h}_{t+1}^{\mathbf{r}}$ is the combined representation processed by a deep decoder $\mathbf{D}$ to forecast the image $\mathbf{\hat{u}}_{t+1}$.

Figure~\ref{fig:fig2}(b) shows the "unfolded" PhyDNet. An input video $\mathbf{u}_{1:T} = (\mathbf{u}_1,...,\mathbf{u}_T) \in \mathbb{R}^{T\times n \times m \times c}$ with spatial size $n \times m$ and $c$ channels is projected into $\bm{\mathcal{H}}$ by the
encoder $\mathbf{E}$ and processed by the recurrent block unfolded in time. This forms a  Sequence To Sequence architecture~\cite{sutskever2014sequence} suited for multi-step prediction, outputting $\Delta$ future frame predictions $\mathbf{\hat{u}}_{T+1:T+\Delta}$. Encoder, decoder and recurrent block parameters are all trained end-to-end, meaning that PhyDNet learns itself without supervision the latent space $\bm{\mathcal{H}}$ in which physics and residual factors are disentangled.

\subsection{PhyCell: a deep recurrent physical model}
\label{section:phycell}

PhyCell is a new physical cell, whose dynamics is governed by the PDE response function $\bm{\mathcal{M}}_p(\mathbf{h^p},\mathbf{u})$\footnote{In the sequel, we drop the index $\mathbf{p}$ in $\mathbf{h^p}$ for the sake of simplicity}: 
 \begin{equation}
     \bm{\mathcal{M}}_p(\mathbf{h},\mathbf{u}) := \Phi(\mathbf{h})+ \mathcal{C}(\mathbf{h},\mathbf{u}) 
\label{eq:Mp}
\end{equation}
where $\Phi(\mathbf{h})$ is a physical predictor modeling only the latent dynamics and $C(\mathbf{h},\mathbf{u})$ is a correction term  modeling the interactions between latent state and input data. \vspace{0.25cm} \\
\textbf{Physical predictor:} $\Phi(\mathbf{h})$ in Eq~(\ref{eq:Mp}) is modeled as follows: 
 \begin{equation}
    \Phi(\mathbf{h}(t,\mathbf{x})) = \sum_{i,j: i+j \leq q}  c_{i,j} \dfrac{\partial^{i+j} \mathbf{h}}{\partial x^i \partial y^j}(t,\mathbf{x})
    \label{eq:phi}
\end{equation}
$\Phi(\mathbf{h}(t,\mathbf{x}))$ in Eq~(\ref{eq:phi}) combines the spatial derivatives with coefficients $c_{i,j}$ up to a certain differential order $q$. This generic class of linear PDEs subsumes a wide range of classical physical models, \eg the heat equation, the wave equations, the advection-diffusion equations. \vspace{0.25cm} \\
\textbf{Correction:} $\mathcal{C}(\mathbf{h},\mathbf{u})$ 
in Eq (\ref{eq:Mp}) 
takes the following form: 
\begin{equation}
    \!\!\!\!\!\!\mathcal{C}(\mathbf{h},\mathbf{u}) \!:=\! \mathbf{K}(t,\mathbf{x})\odot \left[\mathbf{E} (\mathbf{u}(t,\mathbf{x})) \!-\! (\mathbf{h}(t,\mathbf{x}) \!+\! \Phi(\mathbf{h}(t,\mathbf{x} ))\right]
    \label{eq:corrcont}
\end{equation}
Eq (\ref{eq:corrcont}) computes is the difference between the latent state after physical motion $\mathbf{h}(t,\mathbf{x}) + \Phi(\mathbf{h}(t,\mathbf{x}))$ and the embedded new observed input $\mathbf{E}(\mathbf{u}(t,\mathbf{x}))$. 
$\mathbf{K}(t,\mathbf{x})$ is a gating  factor, where $\odot$ is the Hadamard product.

\subsubsection{Discrete PhyCell} 
\label{sec:discretephicell}
\begin{figure}
    \centering
    \includegraphics[width=\columnwidth]{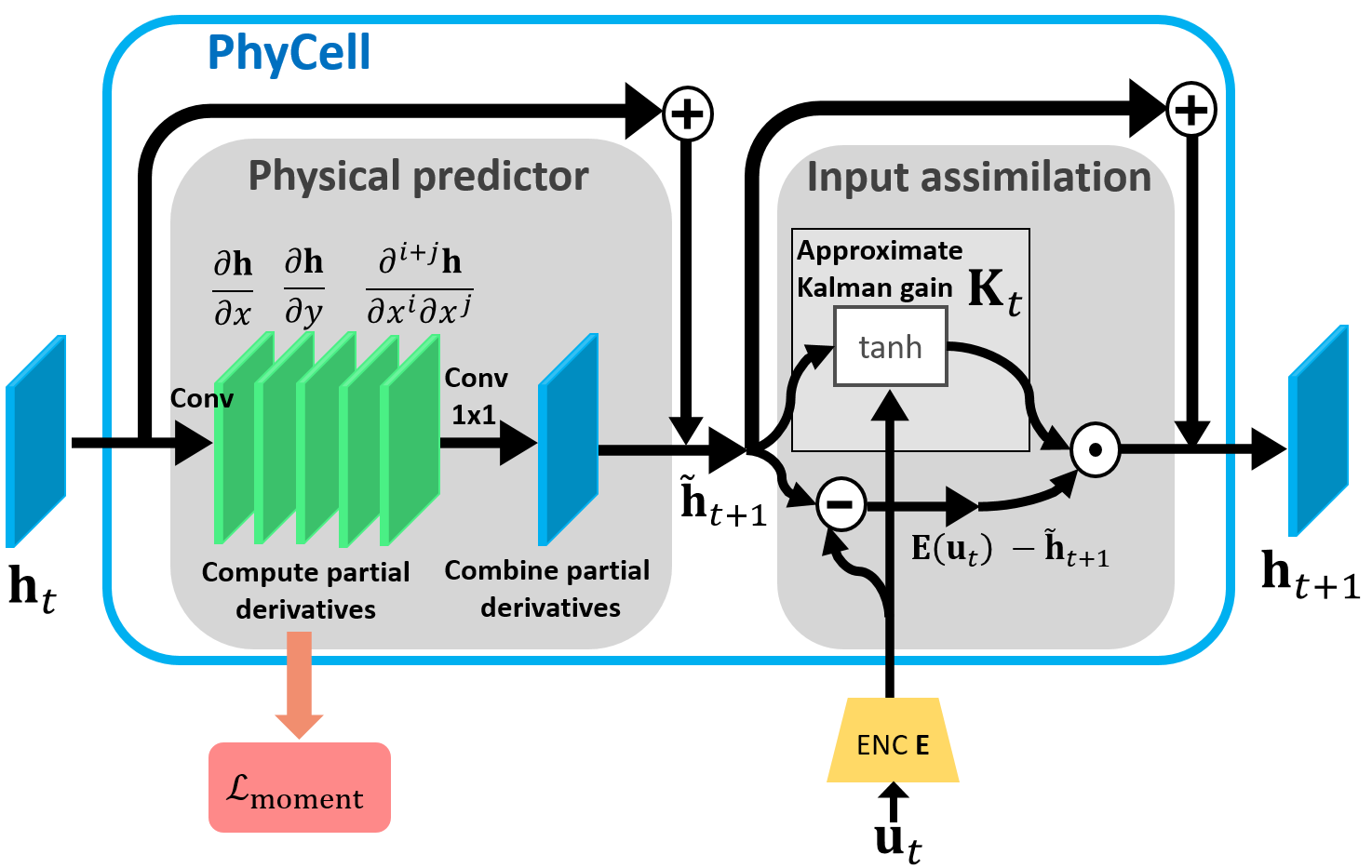}
    \caption{PhyCell recurrent cell implements a two-steps scheme: physical prediction with convolutions for approximating and combining spatial derivatives (Eq (\ref{eq:prediction}) and Eq~(\ref{eq:phi})), and input assimilation as a correction of latent physical dynamics driven by observed data (Eq (\ref{eq:correction})). During training, the filter moment loss in red (Eq~(\ref{eq:lmoment})) enforces the convolutional filters to approximate the desired differential operators.}
    \label{fig:phicell}
\end{figure}

We discretize the  continuous time PDE in Eq (\ref{eq:Mp}) with the standard forward Euler numerical scheme \cite{lu2018beyond}, leading to the discrete time PhyCell (derivation in supplementary 1.1):
\begin{equation}
    \mathbf{h}_{t+1} = (1-\mathbf{K}_t) \odot \left(\mathbf{h}_t + \Phi(\mathbf{h}_t) \right) + \mathbf{K}_t \odot \mathbf{E}(\mathbf{u}_t)
    \label{eq:physical_cell}
\end{equation}
Depicted in Figure \ref{fig:phicell}, PhyCell is an atomic recurrent cell for building physically-constrained RNNs. In our experiments, we use one layer of PhyCell but one can also easily stack several PhyCell layers to build more complex models, as done for stacked RNNs \cite{wang2017predrnn,wang2018predrnn++,wang2019memory}. To gain insight into PhyCell in Eq~(\ref{eq:physical_cell}), we write the equivalent two-steps form:
\begin{empheq}[left=\empheqlbrace]{alignat=2}
&   \tilde{\mathbf{h}}_{t+1} \!= \mathbf{h}_{t} +  \Phi(\mathbf{h}_{t})   &  \!\!\!\quad \text{\small{\textbf{Prediction}\!}} \label{eq:prediction}\\
&   \mathbf{h}_{t+1} \!= \tilde{\mathbf{h}}_{t+1}  + \mathbf{K}_t \odot \left( \mathbf{E}(\mathbf{u}_t) - \tilde{\mathbf{h}}_{t+1} \right) & \!\!\! \quad \text{\small{\textbf{Correction}\!}} \label{eq:correction}
\end{empheq}

The prediction step in Eq~(\ref{eq:prediction}) is a physically-constrained motion in the latent space, computing the intermediate representation $\tilde{\mathbf{h}}_{t+1}$. Eq~(\ref{eq:correction}) is a correction step  incorporating input data. This prediction-correction formulation is reminiscent of the way to combine numerical models with observed data in the data assimilation community \cite{asch2016data,bocquet2019data}, \eg with the Kalman filter \cite{kalman1960new}.
~We show in section \ref{sec:training} that this decoupling between prediction and correction can be leveraged to robustly train our model in  long-term forecasting and missing data contexts. $\mathbf{K}_t$ can be interpreted as the Kalman gain controlling the trade-off between both steps.

\begin{table*}[b]
    \begin{adjustbox}{max width=\textwidth}
    \begin{tabular}{l|lll|lll|lll|lll}
    \Xhline{2\arrayrulewidth}
     \multicolumn{1}{c}{} &  \multicolumn{3}{|c}{\textbf{Moving MNIST}} &  \multicolumn{3}{|c}{\textbf{Traffic BJ}} &  \multicolumn{3}{|c}{\textbf{Sea Surface Temperature}}  &  \multicolumn{3}{|c}{\textbf{Human 3.6}}  \\ 
     \hline
Method  & MSE & MAE & SSIM & MSE $\times 100$  & MAE & SSIM & MSE $\times 10$ & MAE & SSIM  & MSE  / 10 & MAE $/ 100$ & SSIM \\
      \hline \hline

    ConvLSTM \cite{xingjian2015convolutional} & 103.3 & 182.9 & 0.707  & $48.5^*$ & $17.7^*$ & $0.978^*$ & $45.6^*$ & $63.1^*$ & $0.949^*$ & $50.4^*$  & $18.9^*$ & $0.776^*$  \\ 
    PredRNN \cite{wang2017predrnn}  & 56.8 & 126.1 & 0.867 & 46.4 & $17.1^*$ & $0.971^*$ & 41.9 & 62.1  & 0.955 & 48.4 & 18.9 & 0.781 \\ 
    Causal LSTM \cite{wang2018predrnn++}  & 46.5  & 106.8  & 0.898 & 44.8 & $16.9^*$ & $0.977^*$  & $39.1^*$  & $62.3^*$ & $0.929^*$ & 45.8  & 17.2   & 0.851 \\     
    MIM \cite{wang2019memory}  & 44.2  & 101.1 & 0.910 & 42.9 & $16.6^*$ & $0.971^*$ & $42.1^*$ & $60.8^*$  & $0.955^*$  & 42.9 & 17.8 & 0.790 \\
    E3D-LSTM \cite{wang2018eidetic}  & 41.3  & 86.4  & 0.920  & $43.2^*$ & $16.9^*$  & $0.979^*$   & $34.7^*$ & $59.1^*$  & $0.969^*$  & 46.4  & 16.6  & 0.869 \\ \hline
    Advection-diffusion \cite{de2017deep}  & -  &  - &    -& - & - &- &  $34.1^*$ & $54.1^*$ & $0.966^*$ & - & - &-  \\ 
    DDPAE \cite{hsieh2018learning}  & 38.9 & $90.7^*$ & $0.922^*$  & - &- &- &- &- &- &- &- &- \\ \hline
    \textbf{PhyDNet}  & \textbf{24.4} & \textbf{70.3} & \textbf{0.947} & \textbf{41.9} & \textbf{16.2} & \textbf{0.982}    & \textbf{31.9} & \textbf{53.3} & \textbf{0.972} & \textbf{36.9} & \textbf{16.2} & \textbf{0.901} \\ 
    \Xhline{2\arrayrulewidth}
    \end{tabular}
    \end{adjustbox}
    \caption{Quantitative forecasting results of PhyDNet compared to baselines using various datasets. Numbers are copied from original or citing papers. * corresponds to results obtained by running online code from the authors. The first five baseline are general deep models applicable to all datasets, whereas DDPAE \cite{hsieh2018learning} (resp. advection-diffusion flow \cite{de2017deep}) are specific state-of-the-art models for Moving MNIST (resp. SST). Metrics are scaled to be in a similar range across datasets to ease comparison.}
    \label{tab:res1}   
\end{table*}

\subsubsection{PhyCell implementation}
We now specify how the physical predictor $\Phi$ in Eq~(\ref{eq:prediction}) and the correction Kalman gain $\mathbf{K}_t$ in Eq~(\ref{eq:correction}) are implemented.\vspace{0.25cm} \\ 
\textbf{Physical predictor:} we implement $\Phi$ using  a convolutional neural network (left gray box in Figure \ref{fig:phicell}), based on the connection between convolutions and differentiations \cite{dong2017image,long2018pde}.
This offers the possibility to learn a class of filters approximating each partial derivative in Eq~(\ref{eq:phi}), which are constrained by a kernel moment loss, as detailed in section \ref{sec:training}. As noted by~\cite{long2018pde}, the flexibility added by this constrained learning strategy gives better results for solving PDEs than handcrafted derivative filters.
Finally, we use $1 \times 1$ convolutions to linearly combine these derivatives with $c_{i,j}$ coefficients in Eq~(\ref{eq:phi}). \vspace{0.25cm} \\
\textbf{Kalman gain:}
We approximate $\mathbf{K}_t$ in Eq~(\ref{eq:correction}) by a gate with learned  convolution kernels $\mathbf{W}_h$, $\mathbf{W}_u$ and bias $\mathbf{b}_k$:
\begin{equation}
\mathbf{K}_t =  \tanh \left( \mathbf{W}_{h} * \tilde{\mathbf{h}}_{t+1} + \mathbf{W}_{u} * \mathbf{E}(\mathbf{u}_t) + \mathbf{b}_k \right)
\label{eq:kalman_gain}
\end{equation}
Note that if $\mathbf{K}_t = \mathbf{0}$, the input is not accounted for and the dynamics follows the physical predictor 
; if $\mathbf{K}_t = 1 $, the latent dynamics is resetted and only driven  by the input. This is similar to gating mechanisms in LSTMs or GRUs. \vspace{0.25cm} \\ 
\textbf{Discussion:} With specific $\Phi$ predictor,  
$\mathbf{K}_t$ gain and  encoder $\mathbf{E}$, PhyCell recovers recent models from the literature:
\vspace{-0.2cm}
\begin{table}[H]
    \centering
        \begin{adjustbox}{max width=\columnwidth}
    \begin{tabular}{c|ccc}
    model & $\Phi$ & $\mathbf{K}_t$ & $\mathbf{E}$ \\ \hline
    PDE-Net \cite{long2019pde}    &  Eq (\ref{eq:prediction}) & $\mathbf{0}$ & $\mathbf{Id}$     \\ \hline
    Advection-diffusion & advection-diffusion & $\mathbf{0}$ &  $\mathbf{Id}$ \\ 
    flow~\cite{de2017deep}  & predictor & &  \\ \hline
    RKF \cite{becker2019recurrent}  &  locally linear, no  & approx.   & deep encoder   \\
            ~   & phys. constraint     & Kalman gain  &  \\ \hline
            PhyDNet (ours) &  Eq (\ref{eq:prediction})  & Eq (\ref{eq:kalman_gain})  & deep encoder
    \end{tabular}
    \vspace{-0.2cm}
    \label{tab:my_label}
    \end{adjustbox}
\end{table}
\vspace{-0.2cm}
PDE-Net~\cite{long2018pde} directly works on raw pixel data (identity encoder $\mathbf{E}$) and assumes Markovian dynamics (no correction, $\mathbf{K}_t\!\!\!=\!\!\!\mathbf{0}$): the model solves the autonomous PDE $\frac{\partial \mathbf{u}}{\partial t}=\Phi(\mathbf{u})$ given in Eq (\ref{eq:prediction}) but in pixel space.
~This prevents from modeling time-varying PDEs such as those tackled in this work,
~\eg  varying advection terms. 
~The 
flow model in~\cite{de2017deep} uses the closed-form solution of the advection-diffusion equation as predictor ; it is however limited only to this PDE, whereas PhyDNet models a much broader class of PDEs. The Recurent Kalman Filter (RKF)~\cite{becker2019recurrent} also proposes a prediction-correction scheme in a deep latent space, but their approach does not include any prior physical information, and the prediction step is locally linear, whereas we use deep models. An approximated form of the covariance matrix is used for estimating $\mathbf{K}_t$ in~\cite{becker2019recurrent}, which we find experimentally inferior to our gating mechanism in Eq~(\ref{eq:kalman_gain}).

\subsection{Training}
\label{sec:training}

Given a training set of $N$ videos $\bm{\mathcal{D}} =  \left \{ \mathbf{u}^{(i)} \right \} _{i=\{1:N \}}$ and PhyDNet parameters $\mathbf{w}= (\mathbf{w_p},\mathbf{w_r},\mathbf{w_s})$, where $\mathbf{w_p}$ (resp.  $\mathbf{w_r}$) are parameters of the PhyCell (resp. residual) branch,  and $\mathbf{w_s}$ are encoder and decoder shared parameters, we minimize the following objective function:
\begin{equation}
    \mathcal{L}(\bm{\mathcal{D}},\mathbf{w}) = \mathcal{L}_{\text{image}}(\bm{\mathcal{D}},\mathbf{w}) + \lambda  \mathcal{L}_{\text{moment}}(\mathbf{w_p})
\end{equation}
We use the $L^2$ loss for the image reconstruction loss $\mathcal{L}_{\text{image}}$, as commonly done in the literature \cite{wang2017predrnn,wang2018predrnn++,oliu2018folded,wang2018eidetic,wang2019memory}. 

$\mathcal{L}_{\text{moment}}(\mathbf{w_p})$ imposes physical constraints on the $k^2$ learned filters  $  \left\{ \mathbf{w}^k_{p,i,j}\right\}_{i,j \leq k}$, such that each $\mathbf{w}^k_{p,i,j}$ of size $k \times k$ approximates $\frac{\partial^{i+j}}{\partial x^i y^j}$. This is achieved by using a loss based on the moment matrix $\mathbf{M}(\mathbf{w}^k_{p,i,j})$~\cite{long2019pde}, representing the order of the filter differentiation~\cite{dong2017image}.  $\mathbf{M}(\mathbf{w}^k_{p,i,j})$ is compared to a target moment matrix $\mathbf{\Delta}^k_{i,j}$ (see $\mathbf{M}$ and $\mathbf{\Delta}$ computations in supplementary 1.2), leading to: 
 \begin{equation}
   \mathcal{L}_{\text{moment}} = \sum\limits_{i \leq k} \sum\limits_{j \leq k} ||\mathbf{M}(\mathbf{w}^k_{p,i,j}) - \mathbf{\Delta}^k_{i,j} ||_F 
   \label{eq:lmoment}
 \end{equation}

\textbf{Prediction mode} 
~An appealing feature of PhyCell is that we can use and train the model in a "prediction-only" mode by setting $\mathbf{K}_t = \mathbf{0}$ in Eq (\ref{eq:correction}), \ie by only relying on the physical predictor $\Phi$ in Eq (\ref{eq:prediction}). It is worth mentioning that the "prediction-only" mode is not applicable to standard Seq2Seq RNNs: although the decomposition in Eq~(\ref{eq:Mp}) still holds, \ie $\bm{\mathcal{M}}_r(\mathbf{h},\mathbf{u}) = \Phi(\mathbf{h})+ \mathcal{C}(\mathbf{h},\mathbf{u})$, the resulting predictor is naive and useless for multi-step prediction  $\mathbf{\tilde{h}}_{t+1}=0$, see supplementary 1.3).

Therefore, standard RNNs are not equipped to deal with unreliable input data $\mathbf{u}_t$. We show in section~\ref{sec:expe_prediction} that the gain of PhyDNet over those models increases in two important contexts with unreliable inputs: multi-step prediction and dealing with missing data.

\section{Experiments}
\label{section4}

\begin{figure*}[ht]
    \centering
    \includegraphics[width=\textwidth]{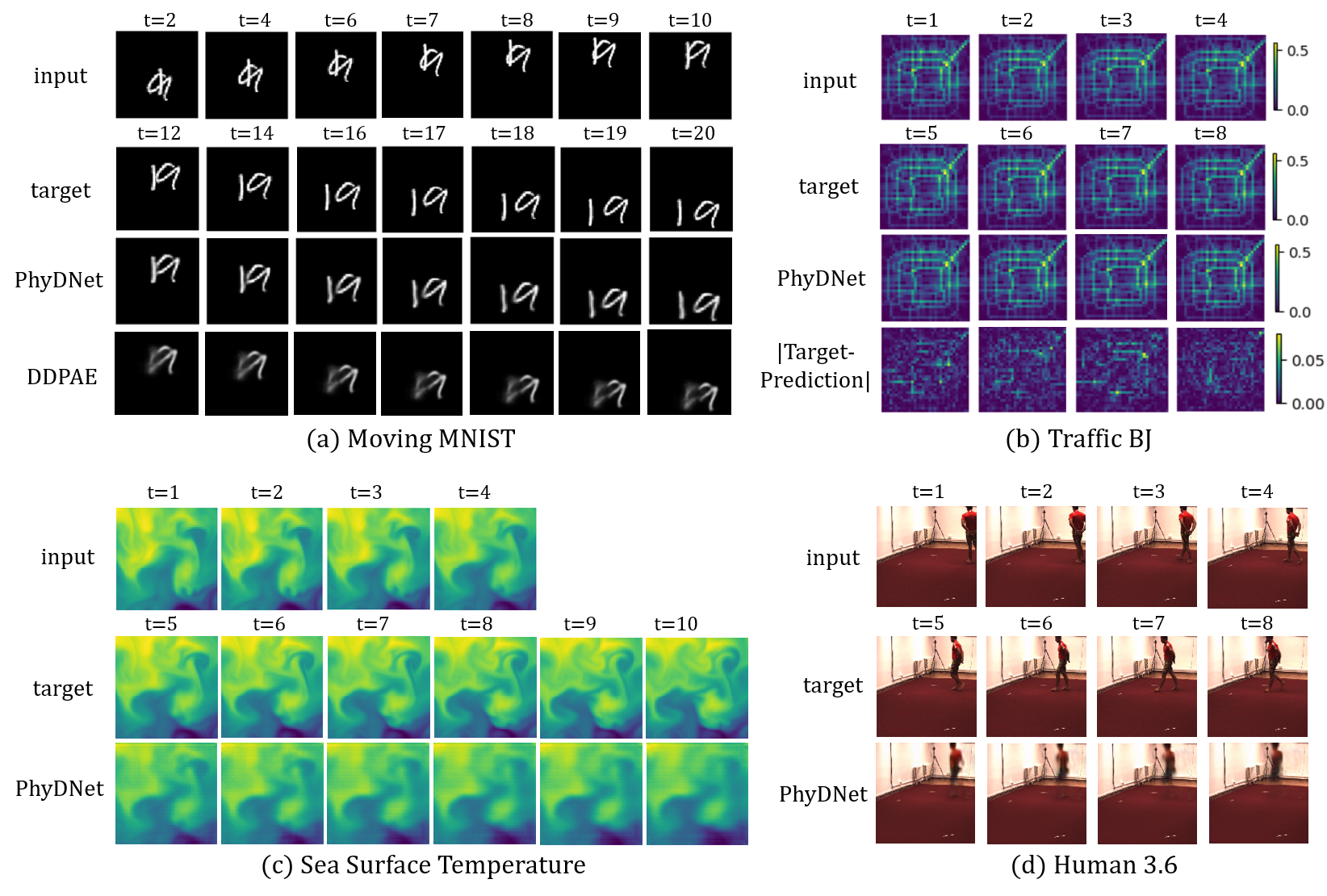}
    \caption{Qualitative results of the predicted frames by PhyDNet for all datasets. First line is the input sequence, second line the target and third line PhyDNet prediction. For Moving MNIST, we add a fourth line with the comparison to DDPAE \cite{hsieh2018learning} and for Traffic BJ the difference $|\text{Prediction-Target}|$ for better visualization.}
    \label{fig:visus}
\end{figure*}

\subsection{Experimental setup}

\paragraph{Datasets}  We evaluate PhyDNet on four datasets from various origins. \textbf{Moving MNIST}~\cite{srivastava2015unsupervised} is a standard synthetic benchmark in video prediction with two random digits bouncing on the walls. \textbf{Traffic BJ} \cite{zhang2017deep} represents complex real-world traffic flows, which requires modeling transport phenomena and traffic diffusion for prediction. \textbf{SST} (Sea Surface Temperature) \cite{de2017deep} consists in meteorological data, whose evolution is governed by the physical laws of fluid dynamics. Finally, \textbf{Human 3.6} \cite{ionescu2013human3} represents general human actions with complex 3D articulated motions. We give details about all datasets in supplementary 2.1.

\paragraph{Network architectures and training}
PhyDNet shares a common backbone architecture for all datasets where the physical branch contains 49 PhyCells ($7 \times 7$ kernel filters) and the residual branch is composed of a 3-layers ConvLSTM with 128 filters in each layer. We set up the trade-off parameter between $\mathcal{L}_{\text{image}}$ and $\mathcal{L}_{\text{moment}}$ to $\lambda=1$. Detailed architectures and $\lambda$ impact are given in supplementary 2.2. Our code is available at \url{https://github.com/vincent-leguen/PhyDNet}.  \vspace{-0.4cm}

\paragraph{Evaluation metrics} We follow evaluation metrics commonly used in state-of-the-art video prediction methods: the Mean Squared Error (MSE), Mean Absolute Error (MAE) and the Structural Similarity (SSIM) \cite{wang2004image} that computes the perceived image quality with respect to a reference. Metrics are averaged for each frame of the output sequence. Lower MSE, MAE and higher SSIM indicate better performances.

\subsection{State of the art comparison}

We evaluate PhyDNet against strong recent baselines, including very competitive data-driven RNN architectures: ConvLSTM  \cite{xingjian2015convolutional}, PredRNN \cite{wang2017predrnn}, Causal LSTM \cite{wang2018predrnn++}, Memory in Memory (MIM) \cite{wang2019memory}. We also compare to methods dedicated to specific datasets: DDPAE \cite{hsieh2018learning}, a disentangling method specialized and state-of-the-art on Moving MNIST ; and the physically-constrained advection-diffusion flow model \cite{de2017deep} that is state-of-the-art for the SST dataset.

\begin{table*}
    \begin{adjustbox}{max width=\textwidth}
    \begin{tabular}{l|lll|lll|lll|lll}
    \Xhline{2\arrayrulewidth}
     \multicolumn{1}{c}{} &  \multicolumn{3}{|c|}{\textbf{Moving MNIST}} &  \multicolumn{3}{|c|}{\textbf{Traffic BJ}} &  \multicolumn{3}{|c|}{\textbf{Sea Surface Temperature}} &  \multicolumn{3}{|c}{\textbf{Human 3.6}}  \\ 
     \hline
    Method  & MSE & MAE & SSIM & MSE $\times$ 100 & MAE & SSIM & MSE $\times$ 10 & MAE & SSIM & MSE $/$ 10 & MAE $/$ 100 & SSIM \\  \hline \hline
    ConvLSTM  & 103.3  & 182.9  & 0.707  & $48.5^*$ & $17.7^*$  & $0.978^*$  &  $45.6^*$ & $63.1^*$ & $0.949^*$  & $50.4^*$ & $18.9^*$  & $0.776^*$ \\
    PhyCell    & 50.8 & 129.3  & 0.870  & 48.9  & 17.9 & 0.978  & 38.2  & 60.2 & 0.969  & 42.5  & 18.3 & 0.891 \\ 
    PhyDNet   & \textbf{24.4}  & \textbf{70.3}  & \textbf{0.947}  &  \textbf{41.9} & \textbf{16.2} & \textbf{0.982}  & \textbf{31.9}  & \textbf{53.3}  & \textbf{0.972}  & \textbf{36.9} & \textbf{16.2} & \textbf{0.901}  \\ 
    \Xhline{2\arrayrulewidth}
    \end{tabular}
    \end{adjustbox}
    \caption{An ablation study shows the consistent performance gain on all datasets of our physically-constrained PhyCell vs the general purpose ConvLSTM, and the additional gain brought up by the disentangling architecture PhyDNet. * corresponds to results obtained by running online code from the authors.}
  \label{tab:ablation}  
\end{table*}

Overall results presented in Table \ref{tab:res1} reveal that PhyDNet outperforms significantly all baselines on all four datasets. The performance gain is large with respect to state-of-the-art general RNN models, with a gain of 17 MSE points for Moving MNIST, 6 MSE points for Human 3.6, 3 MSE points for SST and 1 MSE point for Traffic BJ. In addition, PhyDNet also outperforms specialized models: it gains 14 MSE points compared to the disentangling DDPAE model \cite{hsieh2018learning} specialized for Moving MNIST, and 2 MSE points compared to the advection-diffusion model \cite{de2017deep} dedicated to SST data. PhyDNet also presents large and consistent gains in SSIM, indicating that image quality is greatly improved by the physical regularization. Note that for Human 3.6, a few approaches use specific strategies dedicated to human motion with additional supervision, \eg human pose in \cite{villegas2017learning}. We perform similarly to \cite{villegas2017learning} using only unsupervised training, as shown in supplementary 2.3. This is, to the best of our knowledge, the first time that physically-constrained deep models reach state-of-the-art performances on generalist video prediction datasets.


In Figure \ref{fig:visus}, we provide qualitative prediction results for all datasets, showing that PhyDNet properly forecasts future images for the considered horizons: digits are sharply and accurately predicted for Moving MNIST in (a), the absolute traffic flow error is low and approximately spatially independent in (b), the evolving physical SST phenomena are well anticipated in (c) and the future positions of the person is accurately predicted in (d). We add in Figure \ref{fig:visus}(a) a qualitative comparison to DDPAE \cite{hsieh2018learning}, which fails to predict the future frames properly. Since the two digits overlap in the input sequence, DPPAE is unable to disentangle them. In contrast, PhyDNet successfully learns the physical dynamics of the two digits in a disentangled latent space, leading a correct prediction. In supplementary 2.4, we detail this comparison to DPPAE, and provide additional visualizations for all datasets. 




\subsection{Ablation Study}



We perform here an ablation study to analyse the respective contributions of physical modeling and disentanglement. Results are presented in Table \ref{tab:ablation} for all datasets. We see that a 1-layer PhyCell model (only the left branch of PhyDNet in Figure \ref{fig:fig2}(b)) outperforms a 3-layers ConvLSTM (50 MSE points gained for Moving MNIST, 8 MSE points for Human 3.6, 7 MSE points for SST and equivalent results for Traffic BJ), while PhyCell  has much fewer parameters (270,000 \textit{vs.} 3 million parameters). This confirms that PhyCell is a very effective recurrent cell that successfully incorporates physical prior in deep models.  When we further add our disentangling strategy with the two-branch architecture (PhyDNet), we have another performance gap on all datasets (25 MSE points for Moving MNIST, 7 points for Traffic and SST, and 5 points for Human 3.6), which proves that physical modeling is not sufficient by itself to perform general video prediction and that learning unknown factors is necessary.

\begin{figure}
    \centering
    \includegraphics[width=7.6cm]{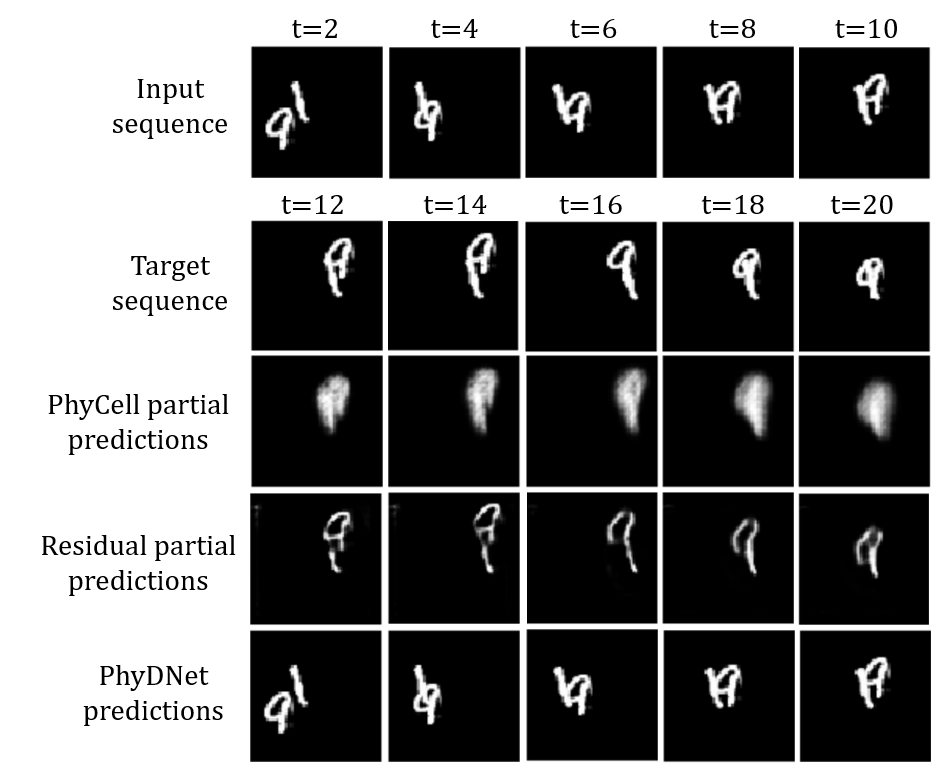}
    \caption{Qualitative ablation results on Moving MNIST: partial predictions show that PhyCell captures coarse localisation of digits, whereas the ConvLSTM branch models the fine shape details of digits. Every two frames are displayed.}
    \label{fig:ablation}
\end{figure}

We qualitatively analyze in Figure~\ref{fig:ablation} partial predictions of PhyDNet for the physical branch  $\hat{\mathbf{u}}^{\mathbf{p}}_{t+1} = \mathbf{D}(\mathbf{h}^{\mathbf{p}}_{t+1})$ and residual branch  $\hat{\mathbf{u}}^{\mathbf{r}}_{t+1} = \mathbf{D}(\mathbf{h}^{\mathbf{r}}_{t+1})$. As noted in Figure \ref{fig:fig1} for Moving MNIST, $\mathbf{h^p}$ captures coarse localisations of objects, while $\mathbf{h^r}$ captures fine-grained details that are not useful for the physical model. Additional visualizations for the other datasets and a discussion on the number of parameters are provided in supplementary 2.5.

\paragraph{Influence of physical regularization}

We conduct in Table \ref{tab:ablation2} a finer ablation on Moving MNIST to study the impact of the physical regularization $\mathcal{L}_{\text{moment}}$ on the performance of PhyCell and PhyDNet. When we disable $\mathcal{L}_{\text{moment}}$ for training PhyCell, performances improve by 7 points in MSE. This underlines that physical laws alone are too restrictive for learning dynamics in a general context, and that complementary factors should be accounted for.  
On the other side, when we disable $\mathcal{L}_{\text{moment}}$ for training our disentangled architecture PhyDNet, performances decrease by 5 MSE points ($29$ \textit{vs} $24.4$) compared to the physically-constrained version. This proves that physical constraints are relevant, but should be incorporated carefully in order to make both branches cooperate. This enables to leverage physical prior, while keeping remaining information necessary for pixel-level prediction. Same conclusions can be drawn for the other datasets, see supplementary 2.6.

\begin{table}[H]
\centering
    \begin{adjustbox}{max width=\columnwidth}
    \begin{tabular}{l|lll}
    \Xhline{2\arrayrulewidth}
     \hline
    Method  & MSE & MAE & SSIM  \\  \hline \hline
    PhyCell & 50.8 & 129.3  & 0.870    \\ 
    PhyCell without $\mathcal{L}_{\text{moment}}$  & 43.4  & 112.8  & 0.895   \\ 
    PhyDNet &  \textbf{24.4}  & \textbf{70.3}  & \textbf{0.947}   \\     
    PhyDNet without $\mathcal{L}_{\text{moment}}$ & 29.0  & 81.2 & 0.934   \\ 
    \Xhline{2\arrayrulewidth}
    \end{tabular}
    \end{adjustbox}
    \caption{Influence of physical regularization for Moving MNIST. }
    \label{tab:ablation2} 
\end{table}

\subsection{PhyCell analysis}
\label{sec:expe_prediction}

\subsubsection{Physical filter analysis}

\begin{figure}[H]
    \centering
    \begin{tabular}{cc}
  
   \hspace{-0.5cm} \includegraphics[height=3cm]{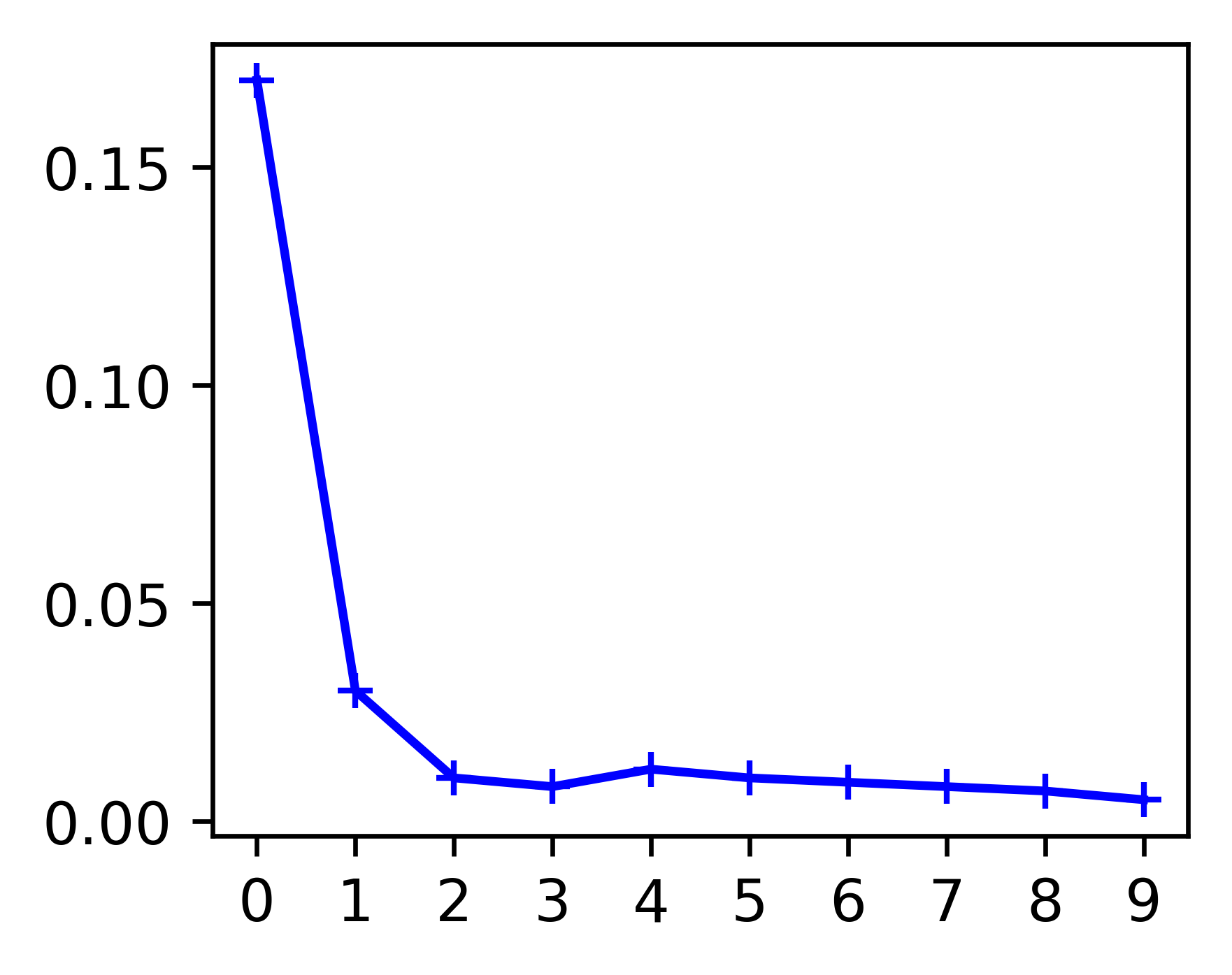}      & \hspace{-0.5cm}
    \includegraphics[height=3cm]{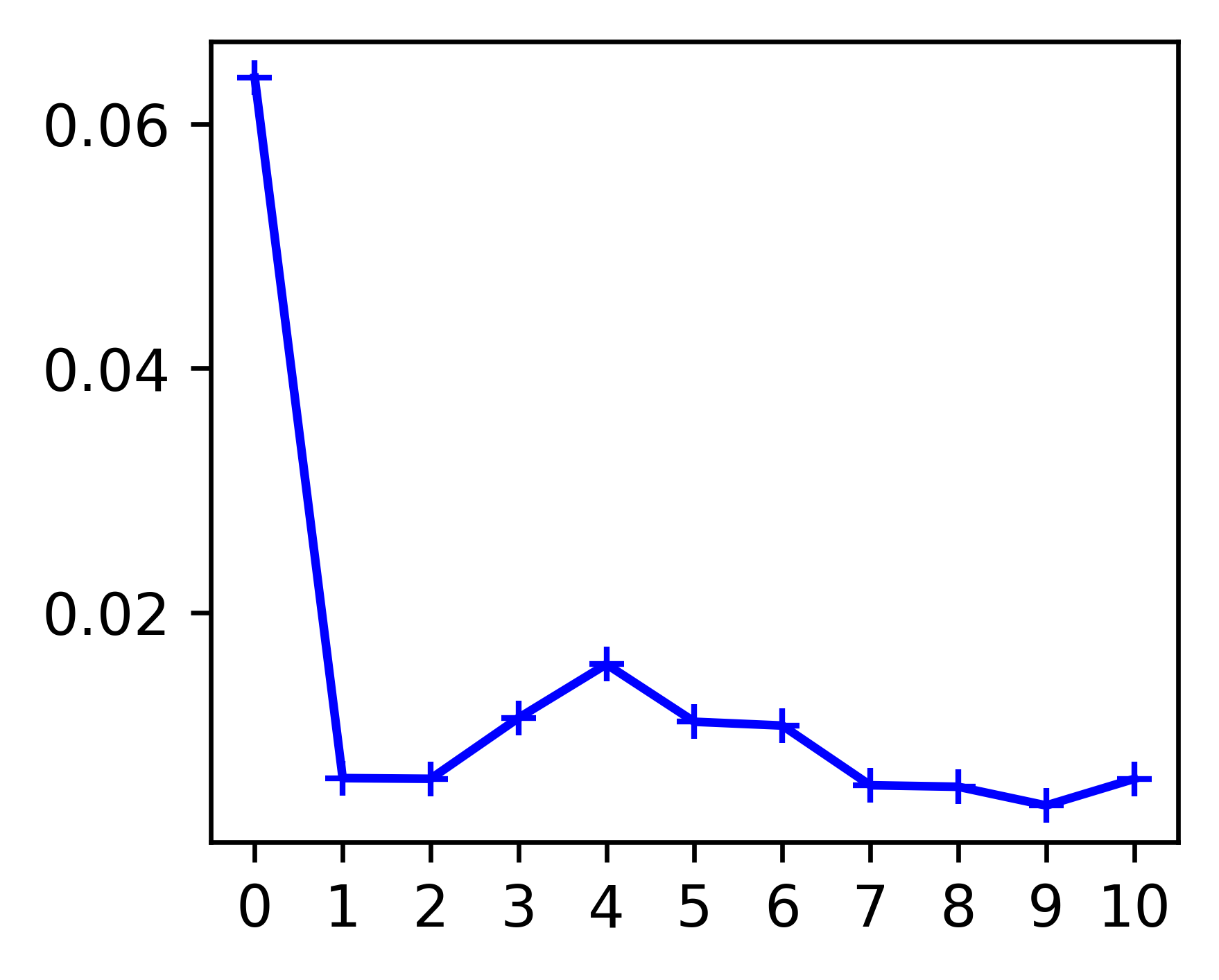}  \\ 
    Moving MNIST & Traffic BJ \\  \hspace{-0.5cm}
    \includegraphics[height=3cm]{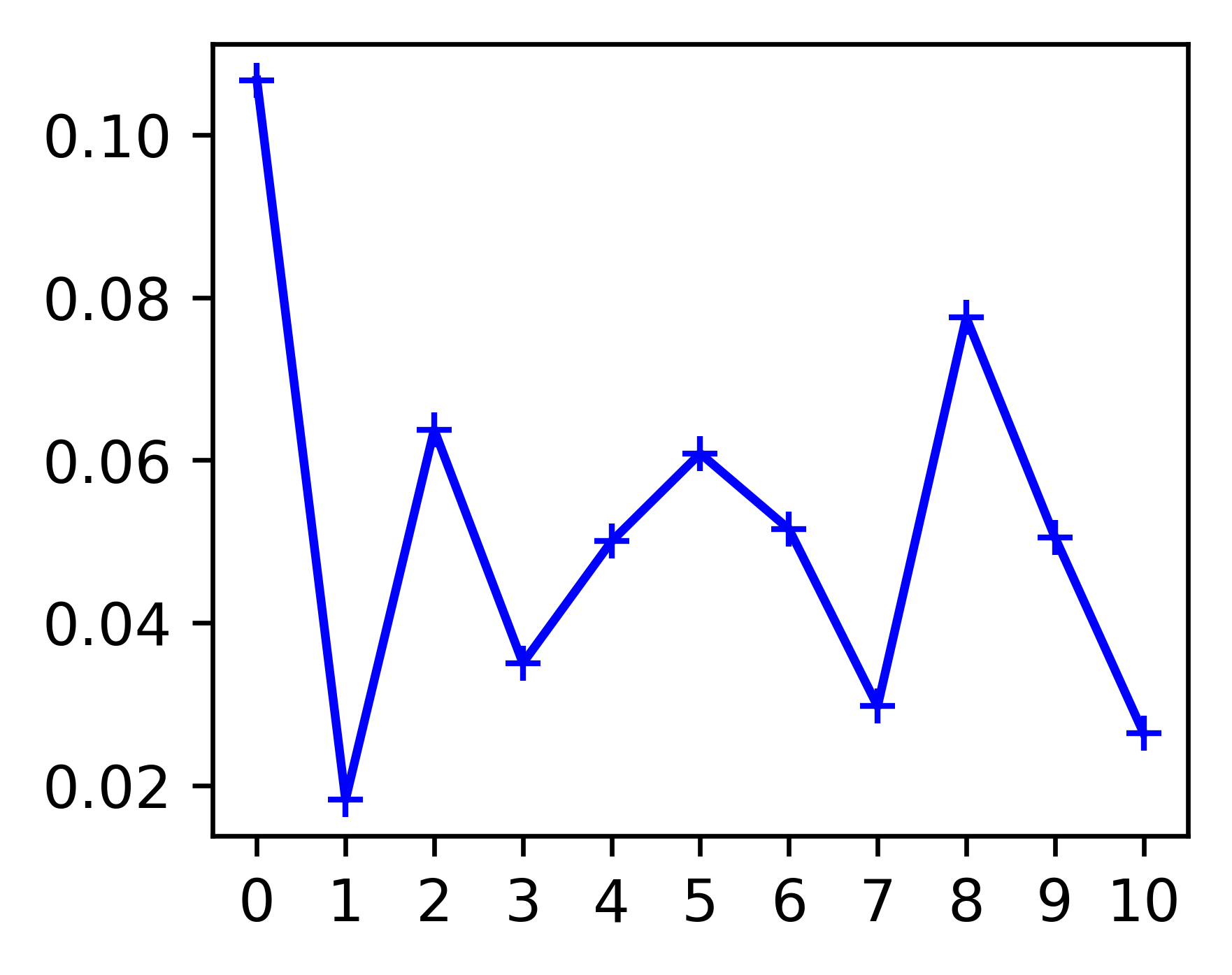}     & \hspace{-0.5cm} \includegraphics[height=3cm]{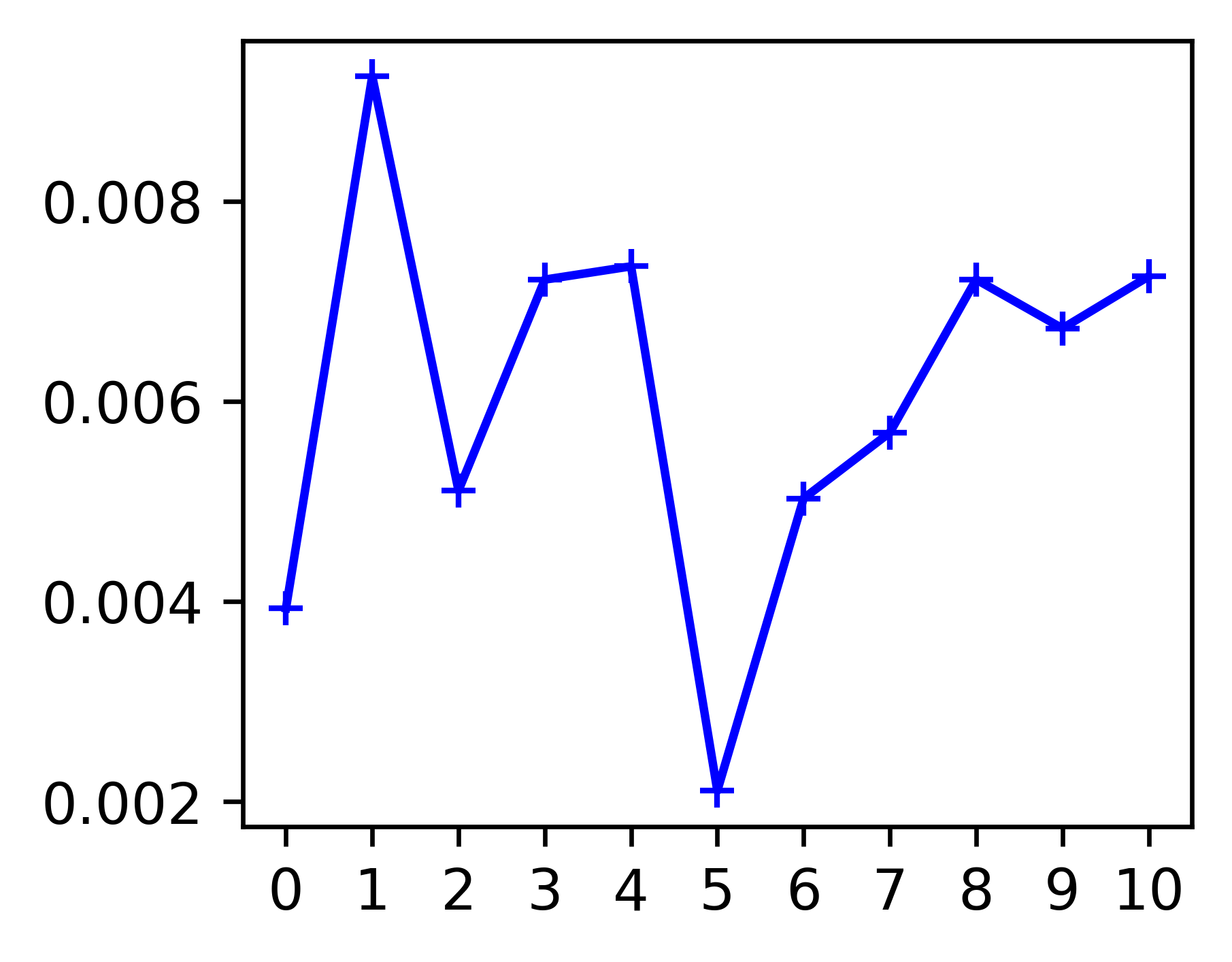} \\
   SST  & Human 3.6

    \end{tabular}{}
    \caption{Mean amplitude of the combining coefficients $c_{i,j}$ with respect to the  order of the differential operators approximated.}
    \label{fig:cij}
\end{figure}

With the same general backbone architecture, PhyDNet can express different PDE dynamics associated to the underlying phenomena by learning specific $c_{i,j}$ coefficients combining the partial derivatives in Eq (\ref{eq:phi}). In Figure \ref{fig:cij}, we display the mean amplitude of the learned coefficients $c_{i,j}$ with respect to the order of differentiation. For Moving MNIST, the $0^{th}$ and $1^{st}$ orders are largely dominant, meaning a purely advective behaviour coherent with the piecewise-constant translation dynamics of the dataset. For Traffic BJ and SST, there is also a global decrease in amplitude with respect to order, we nonetheless notice a few higher order terms appearing to be useful for prediction. 
For Human 3.6, where the nature of the prior motion is less obvious, these coefficients are more spread across order derivatives.
\vspace{-0.2cm}

\vspace{-0.5cm}
\subsubsection{Dealing with unreliable inputs} 
\label{sec:lt-forecasting}

We explore here the robustness of PhyDNet when dealing with unreliable inputs, that can arise in two contexts: long-term forecasting 
and missing data. 
As explained in section~\ref{sec:training}, PhyDNet can be used in a prediction mode in this context, limiting the use of unreliable inputs, whereas general RNNs cannot. To validate the relevance of the prediction mode, we compare PhyDNet to DDPAE \cite{hsieh2018learning}, based on a standard RNN (LSTM) as predictor module. 
Figure \ref{fig:long-term} presents the results in MSE obtained by PhyDNet and DDPAE on Moving MNIST (see supplementary 2.7 for similar results in SSIM).

For long-term forecasting, we evaluate the performances of both methods far beyond the prediction range seen during training (up to 80 frames), as shown in Figure \ref{fig:long-term}(a). We can see that the performance drop (MSE increase rate) is approximately linear for PhyNet, whereas it is much more pronounced for DDPAE. For example, PhyDNet for 80-steps prediction reaches similar performances in MSE than DDPAE for 20-steps prediction. This confirms that PhyDNet can limit error accumulation during forecasting by using a powerful dynamical model.


Finally, we evaluate the robustness of PhyDNet on DDPAE on missing data, by varying the ratio of missing data (from 10 to 50\%) in input sequences during training and testing. 
A missing input image is replaced with a default value (0) image. In this case, PhyCell 
relies only on its latent dynamics by setting $\mathbf{K}_t=0$, whereas DDPAE takes the null image as input. Figure \ref{fig:long-term}(b) shows that the performance gap between PhyDNet and DDPAE increases with the percentage of missing data. 

\vspace{-0.2cm}
\begin{figure}[H]
    \centering
    \begin{tabular}{cc}
    \hspace{-0.8cm} \includegraphics[width=4.5cm]{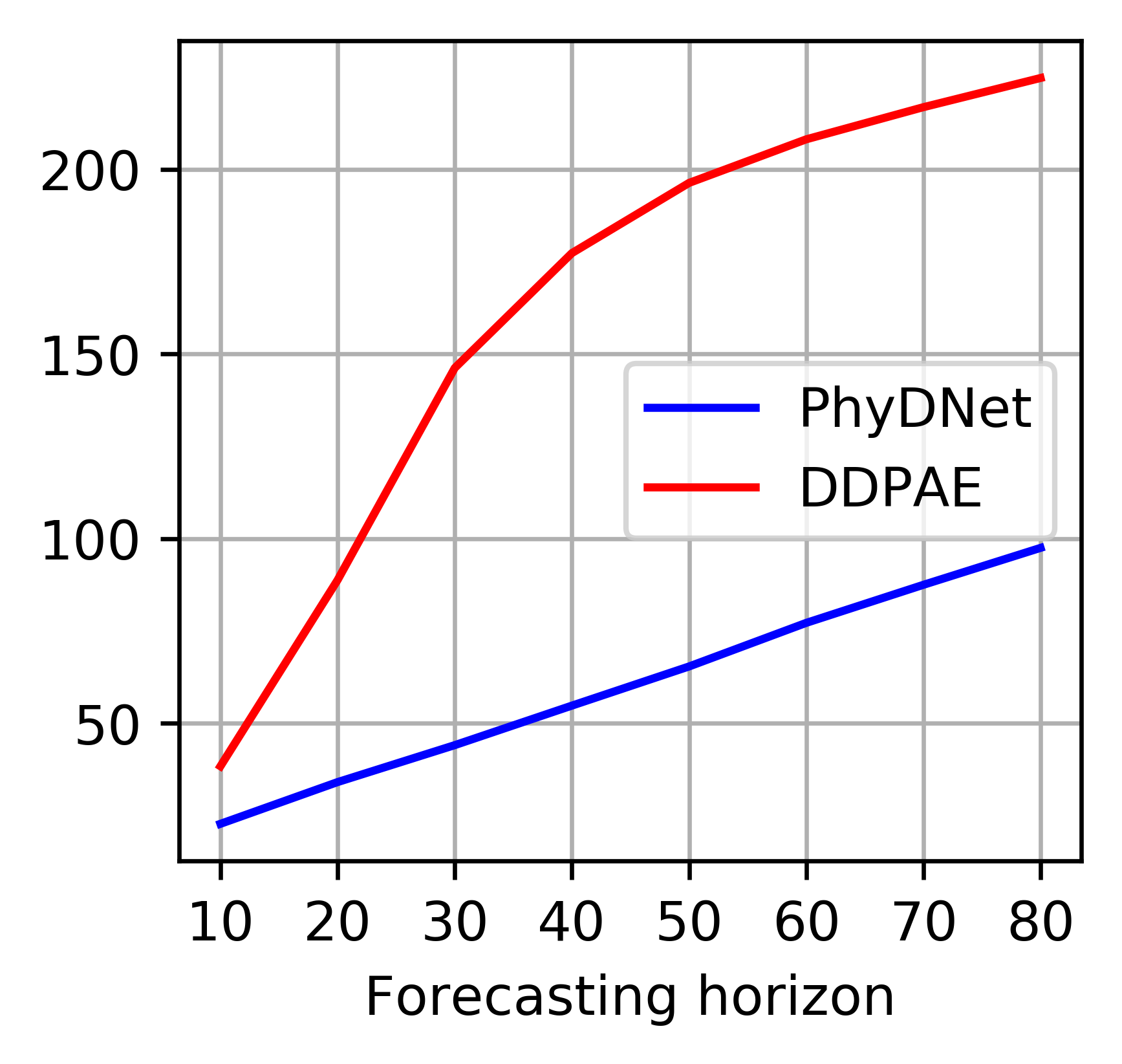}     &  \hspace{-0.5cm} \includegraphics[width=4.5cm]{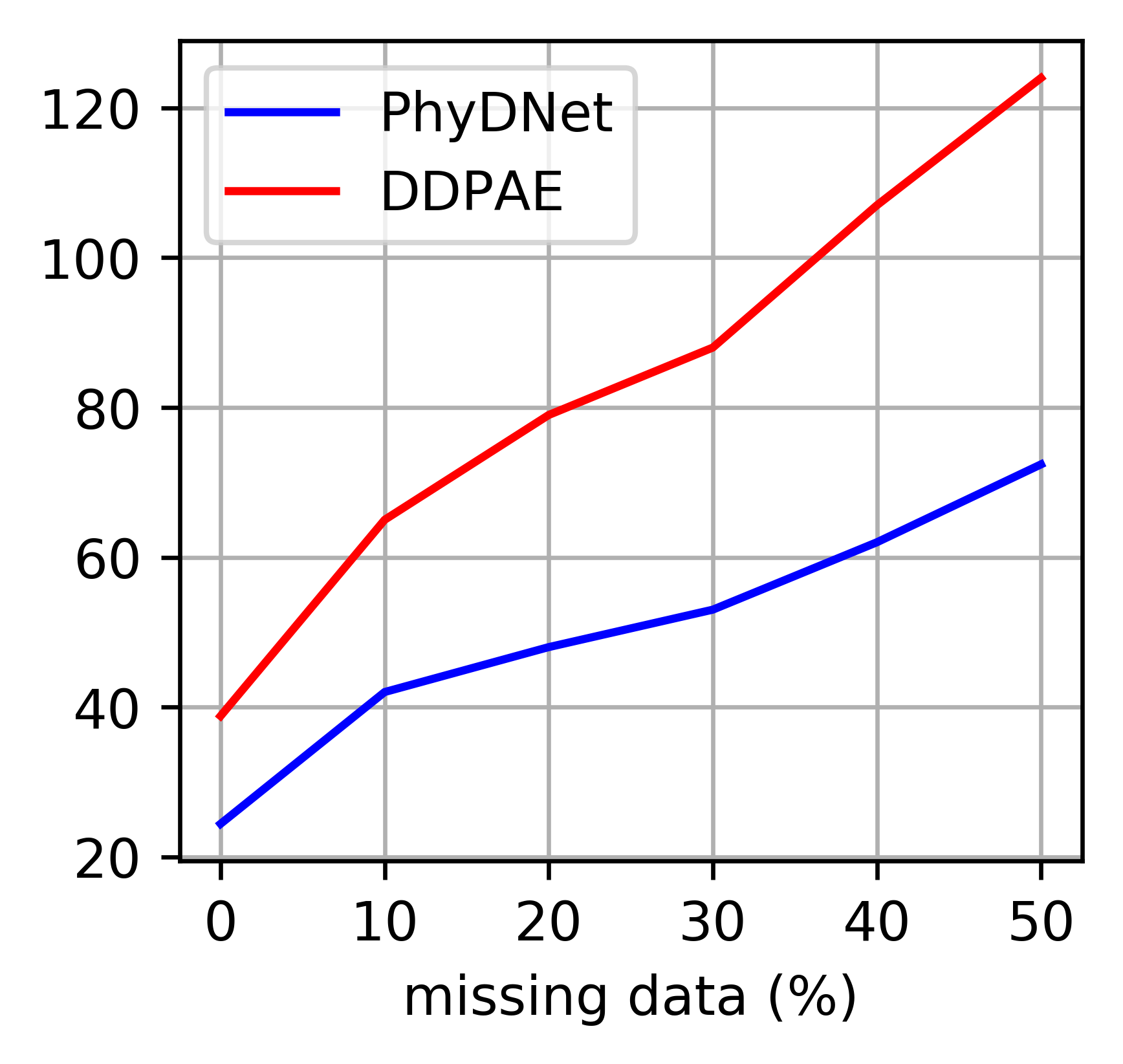} \\
        (a) Long-term forecasting & (b) Missing data 
    \end{tabular} \\

    \caption{MSE comparison between PhyDNet and DDPAE \cite{hsieh2018learning} when dealing with unreliable inputs.}
    \label{fig:long-term}
\end{figure}



\section{Conclusion}

We propose PhyDNet, a new model for disentangling prior dynamical knowledge from other factors of variation required for video prediction. PhyDNet enables to apply PDE-constrained prediction beyond fully observed physical phenomena in pixel space, and to outperform state-of-the-art performances on four generalist datasets. Our introduced recurrent physical cell for modeling PDE dynamics generalizes recent models and offers the appealing property to decouple prediction from correction. Future work include using more complex numerical schemes, \eg Runge-Kutta \cite{fablet2018bilinear}, and extension to probabilistic forecasts with uncertainty estimation \cite{gal2016dropout,corbiere19}, \eg with stochastic differential equations \cite{jia2019neural}.

{\small
\bibliographystyle{ieee}
\bibliography{egbib}
}

\end{document}